\documentclass[10pt,twocolumn,letterpaper]{article}

\usepackage[pagenumbers]{cvpr} 

\usepackage{mmstyle}
\usepackage{graphicx}
\usepackage{amsmath}
\usepackage{amssymb}
\usepackage{xcolor}
\definecolor{Graylight}{gray}{0.9}
\usepackage{colortbl}
\usepackage{booktabs}
\usepackage{multirow}
\usepackage{mathrsfs}
\usepackage{marvosym}
\usepackage{enumerate}
\usepackage[hang,flushmargin]{footmisc} 
\usepackage{tablefootnote}

\usepackage[accsupp]{axessibility} 

\usepackage[pagebackref,breaklinks,colorlinks]{hyperref}

\def\paperID{} 
\def\confName{CVPR}
\def\confYear{2024}

\begin{document}


\title{InternLM-XComposer2: Mastering Free-form Text-Image \\ Composition and Comprehension in Vision-Language Large Models }

\author{Xiaoyi Dong$^{*1,2}$, Pan Zhang$^{*1}$, Yuhang Zang$^{*1}$, Yuhang Cao$^{1,2}$, Bin Wang$^{1}$, Linke Ouyang$^{1}$, Xilin Wei$^{1}$, \\ Songyang Zhang$^{1}$, Haodong Duan$^{1}$, 
Maosong Cao$^{1}$, Wenwei Zhang$^{1}$, Yining Li$^{1}$, Hang Yan$^{1}$, Yang Gao$^{1}$, \\ Xinyue Zhang$^{1}$, Wei Li$^{1}$,
Jingwen Li$^{1}$,
Kai Chen$^{1}$, Conghui He$^{3}$, Xingcheng Zhang$^{3}$, \\ Yu Qiao$^{1}$, Dahua Lin$^{1,2}$, Jiaqi Wang$^{1,}${\textsuperscript{\Letter}}\\
$^1$Shanghai Artificial Intelligence Laboratory,  $^2$The Chinese University of Hong Kong, $^3$SenseTime Group \\
\tt\small
internlm@pjlab.org.cn
}

\twocolumn[{
\renewcommand\twocolumn[1][]{#1}
\maketitle
\begin{center}
    \centering
    \vspace{-20pt}
    \includegraphics[width=1.0\linewidth]{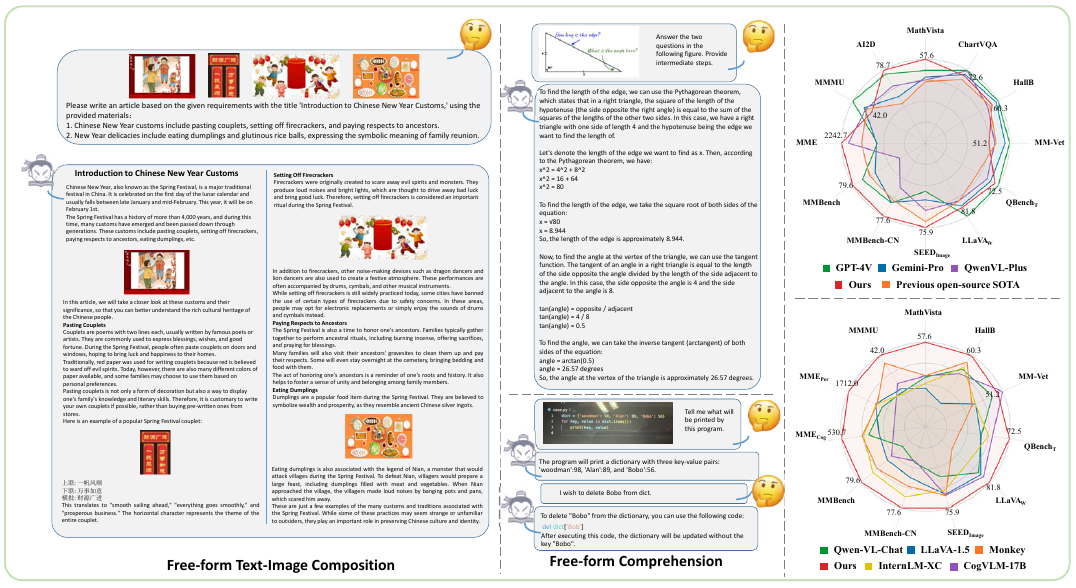}
    \setlength{\abovecaptionskip}{0mm}
    \vspace{-10pt}
    \captionof{figure}{\small
        Overview of free-form text-image composition and comprehension of InternLM-XComposer2.
        Our model based on InternLM2-7B~\cite{2023internlm} not only significantly outperforms existing multimodal models but also \textbf{matches or even surpasses GPT-4V~\cite{openai2023gpt4} and Gemini Pro~\cite{geminiteam2023gemini} in certain assessments}. (Please zoom-in to see the details.)
	}
	\label{fig:teaser}
    \vspace{-5pt}
\end{center}
}]

\maketitle

{\let\thefootnote\relax\footnotetext{\noindent* indicates equal contribution.}}

\begin{abstract}

We introduce InternLM-XComposer2, a cutting-edge vision-language model excelling in free-form text-image composition and comprehension. This model goes beyond conventional vision-language understanding, adeptly crafting interleaved text-image content from diverse inputs like outlines, detailed textual specifications, and reference images, enabling highly customizable content creation. InternLM-XComposer2 proposes a Partial LoRA (PLoRA) approach that applies additional LoRA parameters exclusively to image tokens to preserve the integrity of pre-trained language knowledge, striking a balance between precise vision understanding and text composition with literary talent. Experimental results demonstrate the superiority of InternLM-XComposer2 based on InternLM2-7B in producing high-quality long-text multi-modal content and its exceptional vision-language understanding performance across various benchmarks, where it not only significantly outperforms existing multimodal models but also \textbf{matches or even surpasses GPT-4V and Gemini Pro in certain assessments}. 
This highlights its remarkable proficiency in the realm of multimodal understanding.
The InternLM-XComposer2 model series with 7B parameters are publicly available at \url{https://github.com/InternLM/InternLM-XComposer}.

\end{abstract}
\section{Introduction}
\label{sec:intro}
In recent years, there has been a remarkable evolution in the field of large language models (LLMs)~\cite{raffel2020exploring,brown2020language,chowdhery2022palm,openai2020chatgpt,openai2023gpt4,vicuna2023,touvron2023llama2}. Foremost among these, models like ChatGPT~\cite{openai2020chatgpt} have completely altered human interaction with technology. Concurrently, a variety of open-source LLMs, such as Llama~\cite{touvron2023llama}, Mistra~\cite{jiang2023mistral}, InternLM~\cite{2023internlm}, QWen~\cite{qwen7b}, GLM~\cite{du2022glm}, and Baichuan~\cite{baichuan2023baichuan2}, have empowered the customization of LLMs. Building on these open-source foundations, the community has seen substantial progress in multimodal large language models (MLLMs)~\cite{zhang2023internlm,zhu2023minigpt,liu2023visual,Gao2023LLaMAAdapterVP,dai2023instructblip,llava1_5,wang2023cogvlm,bai2023qwen}. These MLLMs are adept at interpreting images and engaging in text-image dialogues, showcasing impressive multimodal understanding. Unlike traditional MLLMs, a recent innovation, \ie, InternLM-XComposer~\cite{zhang2023internlm}, has focused on using MLLMs for text-image composition and comprehension, marking a novel direction in MLLM research. However, this pioneering work is currently limited to generating text-image articles based on titles alone, lacking the sophistication to meet more complex composition requirements. 
Furthermore, while achieving leading performance at its inception, this model still possesses significant potential for enhancement in detailed perception and complex reasoning capabilities to advance its vision-language comprehension performance.

This observation motivates the development of more advanced vision-language models capable of practical and potent text-image composition and comprehension. In this paper, we introduce InternLM-XComposer2, a cutting-edge model excelling in free-form text-image composition and comprehension, built based on InternLM2~\cite{2023internlm}. InternLM-XComposer2 represents a significant advancement over its predecessor, InternLM-XComposer~\cite{zhang2023internlm}, in both text-image composition and comprehension. InternLM-XComposer2 is adept at producing high-quality, integrated text-image articles from a variety of free-form inputs, such as detailed specifications, structured outlines, and reference images, serving to a wide range of application contexts. In the realm of multimodal understanding, it demonstrates exceptional capabilities in detailed perception, logical reasoning, and extensive knowledge integration. Its performance significantly surpasses that of existing open-source MLLMs, and it stands on par with, or even exceeds, advanced models like GPT-4V~\cite{openai2023gpt4} and Gemini Pro~\cite{geminiteam2023gemini} in various benchmarks.

The appealing capabilities of InternLM-XComposer2 are primarily due to two critical design elements. (1) \textbf{Partial LoRA}: The Partial LoRA (P-LoRA) design harmonizes its abilities in composition and comprehension. This involves feeding forward image tokens with additional LoRA~\cite{hu2022lora} (Low-Rank Adaptation) parameters, while language tokens retain the original architecture. This selective enhancement ensures robust performance in both visual and textual domains. (2) \textbf{High-quality and Diverse Data Foundataion}: The quality and diversity of the training data are pivotal. Our dataset for free-form text-image composition excels in: adhering to complex instructions, customization with text and image for tailored content, high-quality and stylistically diverse writing, and versatile text editing including condensing, expanding, and revising.
For exceptional vision-language comprehension capabilities, we gather a wide range of high-quality pretraining and supervised fine-tuning multimodal data. This collection spans various aspects and types, such as captions, general QA, scientific QA, chat-based QA, mathematical QA, concept knowledge, conversation, and text-image composition.

InternLM-XComposer2 surpasses existing benchmarks in both composition and comprehension. In the creation benchmark of OpenCompass~\cite{2023opencompass} for evaluating the creativity of LLMs, InternLM-XComposer2 showcases outstanding performance. To demostrate our multimodal comphrension capility, we compare our InternLM-XComposer2 on a list of benchmarks with both open-source MLLMs and closed-source APIs, \eg, GPT4V~\cite{openai2023gpt4}, Gemini Pro~\cite{geminiteam2023gemini}, and Qwen-VL Plus~\cite{qwen-vl-plus}. We report results in MathVista~\cite{lu2024mathvista}, MMMU~\cite{yue2023mmmu}, AI2D~\cite{kembhavi2016diagram}, MME~\cite{fu2023mme}, MMBench~\cite{MMBench}, MMBench-Chinese~\cite{MMBench}, SEED-Bench (Image)~\cite{li2023seedbench}, LLaVA-Bench (In-the-Wild)~\cite{liu2023visual}, QBench~\cite{wu2023q}, MM-Vet ~\cite{yu2023mmvet}, HallusionBench~\cite{guan2023hallusionbench}, ChartQA~\cite{masry2022chartqa}, and POPE~\cite{li2023evaluating}. InternLM-XComposer2 based on InternLM2-7B significantly exceeds the performance of existing open-source models by an impressive margin. Remarkably, it demonstrates superior performance to GPT4V~\cite{openai2023gpt4}, Gemini Pro~\cite{geminiteam2023gemini} across six benchmarks.

\section{Related Works}
\label{sec:related}

\noindent{\textbf{Large Language Models (LLMs).}}
Recent LLM architectures have marked a transition from encoder-decoder frameworks (\eg, BERT~\cite{devlin2018bert}, T5~\cite{raffel2020exploring}) to an emphasis on decoder-only models employed with autoregressive training techniques for next-token prediction (\eg, GPT~\cite{radford2018improving}).
The following works (\eg, GPT3~\cite{brown2020language}, InstructGPT~\cite{ouyang2022training}, ChatGPT~\cite{openai2020chatgpt}, PaLM~\cite{chowdhery2022palm}) have seen the integration of advanced techniques such as instruction-tuning and Reinforcement Learning from Human Feedback (RLHF). Coupled with expansive parameter sizes and extensive training data, these LLM models have achieved substantial performance enhancements across a diverse range of Natural Language Processing (NLP) tasks.
Other notable LLMs encompass a range of developments, such as the OPT~\cite{zhang2022opt}, LLaMA series~\cite{touvron2023llama,touvron2023llama2}, \eg, Mistral~\cite{jiang2023mistral,jiang2024mixtral}, InternLM~\cite{2023internlm}, GLM series~\cite{du2022glm,zeng2023glm-130b}, Qwen series~\cite{qwen7b,bai2023qwen}, Baichuan~\cite{baichuan2023baichuan2}, Skywork~\cite{wei2023skywork} and Falcon~\cite{penedo2023refinedweb} have made significant contributions to the field.

\noindent{\textbf{Multimodal Large Language Models (MLLMs).}}
Vision-language models (VLMs), exemplified by CLIP~\cite{radford2021learning} and its subsequent works~\cite{li2022blip,jia2021scaling,fang2023eva,li2021grounded,zhang2022glipv2,liu2023grounding,sun2023alpha}, align image and text features in a unified embedding space. This alignment is achieved through contrastive learning objectives applied to extensive image-text pair datasets. VLMs achieve strong zero-shot and few-shot performance, showcasing significant generalization abilities across a range of downstream tasks.

Benefiting from existing large language models and VLMs as the visual encoder, recent Multimodal Large Language Models~(MLLMs)~\cite{openai2023gpt4,chen2023pali,chen2023palix,chen2023pali3,driess2023palme,fu2023gemini} achieve visual perception, understanding and reasoning abilities, show superb performance in diverse vision-language tasks.
A series of studies~\cite{zhu2023minigpt,dai2023instructblip,liu2023visual,wang2023vigc,zhao2023mllm,zhao2023mmicl,li2023otter,zang2023contextual,chen2023shikra,peng2023kosmos,ye2023mplug,awadalla2023openflamingo,alayrac2022flamingo,chen2023minigptv2,qi2023gemini,li2023monkey,2023xtuner} have explored further improve the MLLM in different dimensions, such as instruction tuning~\cite{liu2023visual,zhao2023mllm,chen2023sharegpt4v}, efficient fine-tuning~\cite{hu2022lora}, high-resolution image inputs~\cite{bai2023qwen,wang2023cogvlm,wei2023vary}, hallucination mitigation~\cite{zhao2023beyond,huang2023opera,yin2023woodpecker}, image generation~\cite{emu_2023,seed_2023,CM3Leon,dong2023dreamllm}, 3D understanding~\cite{qi2023gpt4point} and image-text comprehension and composition~\cite{zhang2023internlm}.

To enable highly customizable content creation, our model is designed for free-form text-image composition and comprehension based on MLLMs.
We use Intern-LM2 as the LLM and CLIP ViT-Large as the visual encoder and propose a new partial LoRA to align the text-image modalities.
Given flexible and multi-modal user inputs such as specifications, outlines, and reference images, our model is capable of generating high-quality interleaved text-image written content.

\section{Method}

\begin{table*}[t]
\centering
\footnotesize
\setlength{\tabcolsep}{5mm}{
\begin{tabular}{ll}
\toprule
Task &  Dataset\\
\midrule
General Semantic Alignment  &  ShareGPT4V-PT~\cite{chen2023sharegpt4v}, COCO~\cite{chen2015microsoft},  Nocaps~\cite{agrawal2019nocaps}, TextCaps~\cite{sidorov2020textcaps},  LAION400M~\cite{schuhmann2021laion},  SBU~\cite{Ordonez_2011_im2text},   CC 3M~\cite{sharma2018conceptual} \\
World Knowledge Alignment & Concept Data~\cite{zhang2023internlm} \\
Vision Capability Enhancement & WanJuan~\cite{He2023WanJuanAC}, Flicker\cite{young2014flicker}, MMC-Instruction\cite{liu2023mmc} \\
\bottomrule

\end{tabular}}
\vspace{-2mm}
\caption {Datasets used for Pre-Training. The data are collected from diverse sources for the three objectives.}
\label{tab:pretrain_data}
\end{table*}

\subsection{Model Architecture}
Our proposed model, InternLM-XComposer2, incorporates a vision encoder and a Language Learning Model (LLM). These two components are interconnected via an innovative Partial LoRA module. Given a set of images and text, the LLM utilizes the output from the vision encoder as visual tokens and the tokenized text as language tokens. These tokens are then concatenated to form the input sequence.

\noindent\textbf{Vision Encoder.}
The vision encoder in our model is designed to extract high-level visual features from raw images. It is pretrained in an image-language contrastive manner(CLIP). Our findings indicate that, when used in conjunction with our Partial LoRA module, a lightweight vision model performs effectively. For the sake of efficiency, we have opted to use the OpenAI ViT-Large model.

\begin{figure}[t!]
	\centering
	\includegraphics[width=0.9\linewidth]{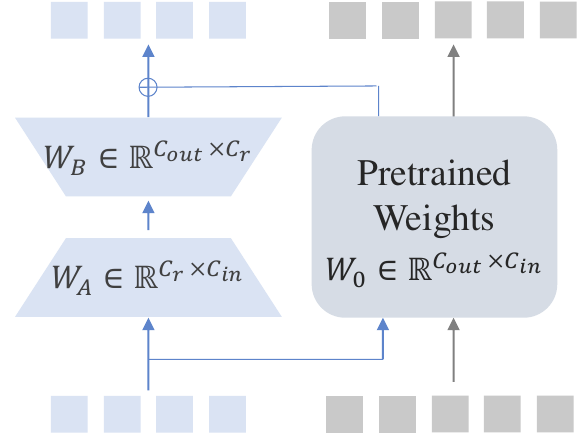}
	\caption{\textbf{The illustration of the Partial-LoRA.} The blue tokens represent the visual tokens and the gray tokens are the language tokens. Our Partial-LoRA is only applied to the visual tokens. }
	\label{fig:plora}
 \vspace{-10pt}
\end{figure}

\noindent\textbf{Large Language Model.}
We employ the recently introduced InternLM-2 as our Large Language Model (LLM). This model boasts exceptional multi-lingual capabilities and has demonstrated impressive results in benchmarks. In practical applications, we utilize the InternLM2-7B-Chat-SFT variant as our LLM.

\noindent\textbf{Partial Low-Rank Adaptation.}
In the realm of multi-modal Language Learning Models (LLMs), one insufficiently explored area is the effective alignment of different modalities. A desired alignment should potentially enrich the LLM with new modality-specific knowledge, while simultaneously preserving its inherent capabilities. Current methods predominantly adopt one of two approaches: they either treat the visual token and language token equally or as entirely distinct entities. We contend that the first approach overlooks the inherent property distinctions between modalities, while the second approach results in a substantial alignment cost.

In our pursuit of effective modality alignment, we introduce Partial LoRA, a versatile plug-in module designed to align knowledge from a new modality to the LLM. As illustrated in Figure X, Partial LoRA draws inspiration from the original LoRA and incorporates a low-rank adaptation that is exclusively applied to the new modality portion of the input tokens. In our specific configuration, Partial LoRA is applied to all visual tokens.

Formally, for each linear layer $L_{0}$ in the LLM blocks, we denote its weight matrix $W_{0} \in \mathbb{R}^{( C_{out}\times C_{in})}$ and bias $B_{0}\in \mathbb{R}^{C_{out}}$, where $C_{in}$ and $C_{out}$ are the input and output dimension. Its corresponding Parital LoRA contains two low-rank matrix $W_{A} \in \mathbb{R}^{ C_{r}\times C_{in}}$ and $W_{B} \in \mathbb{R}^{ C_{out}\times C_{r}}$.
With a given input $x = [x_{v}, x_{t}] $, we have the output feature $\hat{x}$ by:
\begin{align*}
    &\hat{x}_{t} = W_{0}x_{t} + B_0 \\
    &\hat{x}_{v} = W_{0}x_{v} + W_BW_Ax_{v} + B_0 \\
    &\hat{x} = [\hat{x}_{v} , \hat{x}_{t}]     
\end{align*}   
where $x_{v}$ and $x_{t}$ are the visual tokens and language tokens of the input sequence respectively.

\subsection{Pre-Training}
During the pre-training phase, the LLM remains constant while both the vision encoder and Partial LoRA are fine-tuned to align the visual tokens with the LLM. The pre-training data is meticulously curated with \textbf{three objectives} in mind:  1) general semantic alignment, 2) world knowledge alignment, 3) vision capability enhancement. 

\noindent{\textbf{General Semantic Alignment.}}
The objective of general semantic alignment is to equip the MLLM with the fundamental ability to comprehend image content. For instance, the MLLM should be able to recognize that a picture of Einstein represents ‘a human’. We utilize image caption data from a variety of sources for this purpose, including high-quality, detailed captions from ShareGPT4V-PT, as well as concise and precise captions from COCO, NoCaps, TextCaps, \etc. During the pre-training phase, we employ a simple instruction: \textit{Describe this image briefly/in detail}.

\noindent{\textbf{World Knowledge Alignment.}}
World knowledge represents an advanced capability of the MLLM. For instance, the MLLM should be able to identify the man in the figure mentioned above as ‘Albert Einstein’ and further talk something about him. To align the world knowledge depicted in the image with the knowledge already acquired by the LLM, we have constructed a concept dataset. This dataset is carefully filtered from the concept data utilized in InternLM-XComposer~\cite{zhang2023internlm}. Given that the text in the concept data only partially describes the content in the image and their relationship is complex to model, we employ a more broad instruction: \textit{Tell me something about this image}.

\noindent{\textbf{Vision Capability Enhancement.}}
Finally, an advanced MLLM necessitates certain vision-specific capabilities, such as Optical Character Recognition (OCR), object localization (grounding), and the understanding of structured images (\eg, charts, tables). To achieve this, we have compiled relevant datasets, as outlined in Table.\ref{tab:pretrain_data}, and have implemented corresponding instructions for training.

Thanks to the design of Partial LoRA, the LLM is able to adapt to visual tokens while maintaining its original language processing capabilities. The fixed LLM also enables us to directly use in-context learning performance as a measure of pre-training quality.

In our implementation, we employ the OpenAI CLIP ViT-L-14-336 as the vision encoder. We increase its resolution to $490 \times 490$ for improved performance. 
For the Partial LoRA, we set a rank of $256$ for all the linear layers in the LLM decoder block. 
Our training process involves a batch size of 4906 and spans across 2 epochs. The learning rate is initially set to increase to $2 \times 10^{-4}$ within the first $1\%$ of the training steps. Following this, it decreases to $0$ according to a cosine decay strategy. 
To preserve the pre-existing knowledge of the vision encoder, we apply a layer-wise learning rate (LLDR) decay strategy and the decay factor is set to $0.90$.

\begin{table}[t]
\centering
\footnotesize
\setlength{\tabcolsep}{2mm}{
\begin{tabular}{ll}
\toprule
Task &  Dataset\\
\midrule
\multicolumn{2}{l}{\textit{Multi-task training}} \\
Caption  &  ShareGPT4V~\cite{chen2023sharegpt4v}, COCO~\cite{chen2015microsoft},Nocaps~\cite{agrawal2019nocaps} \\
General QA  & VQAv2~\cite{VQAv2}, GQA~\cite{hudson2018gqa}, OK-VQA~\cite{marino2019ok} \\
Science QA  & AI2D~\cite{kembhavi2016diagram}, SQA~\cite{lu2022learn} \\
Chart QA    & DVQA~\cite{kafle2018dvqa}, ChartQA~\cite{masry2022chartqa} \\
Math QA     & MathQA~\cite{amini2019mathqa}, Geometry3K\cite{Geometry3K} \\
World Knowledge QA & A-OKVQA~\cite{schwenk2022okvqa}, ,KVQA~\cite{shah2019kvqa} \\
Conversation & LLaVA-150k~\cite{liu2023visual}, LVIS-Instruct4V~\cite{wang2023see} \\
\midrule
\multicolumn{2}{l}{\textit{Instruction tuning}} \\
Free-from Composiiton  &  In-house data (Refer to Sec.\ref{sec:compos_data})\\
Conversation  &  LLaVA-150k~\cite{liu2023visual}, LVIS-Instruct4V~\cite{wang2023see} \\
& ShareGPT-en\&zh ~\cite{vicuna2023}, InternLM-Chat\cite{2023internlm} \\
\bottomrule

\end{tabular}}
\vspace{-2mm}
\caption {Datasets used for Supervised Fine-Tuning. We collect data from diverse sources to empower the model with different capabilities.}
\label{tab:sft data}
\end{table}

\begin{table*}[t!]
\footnotesize
\centering
\setlength{\tabcolsep}{1.3mm}{
\begin{tabular}{l|cccccccccccc}
\toprule

Method & MathVista & AI2D & MMMU & MME & MMB &MMB$^{CN}$ & SEED$^{I}$ & LLaVA$^{W}$ &   QBench$^{T}$ & MM-Vet & HallB & ChartVQA  \\ \midrule 
Open-Source  & \textit{SPH-MOE} & \textit{Monkey} & \textit{Yi-VL} & \textit{WeMM} & \textit{L-Int2} & \textit{L-Int2} & \textit{SPH-2} & \textit{CogVLM} & \textit{Int-XC} & \textit{CogVLM} & \textit{Monkey} & \textit{CogAgent}  \\ 
Previous SOTA  & 8x7B & 10B & 34B & 6B & 20B & 20B & 17B & 17B & 8B & 30B & 10B & 18B  \\ 
  & 42.3 & 72.6 & 45.9 & \underline{2066.6} & 75.1 & 73.7 & \underline{74.8} & 73.9 & 64.4 & 56.8 & 58.4 & 68.4  \\  
 \midrule \multicolumn{11}{l}{\textit{Closed-source API}} \\
        GPT-4V & \underline{49.9} & \underline{78.2} & \textbf{56.8} & 1926.5 & \underline{77.0} & \underline{74.4} & 69.1 &\textbf{ 93.1 }& \textbf{74.1} & \textbf{67.7} & \textbf{65.8 }& \textbf{78.5}  \\ 
        Gemini-Pro & 45.2 & 73.9 & \underline{47.9} & 1933.3 & 73.6 & 74.3 & 70.7 & 79.9 & 70.6 & \underline{64.3} & \underline{63.9} & 74.1  \\ 
        QwenVL-Plus & 43.3 & 75.9 & 46.5 & 2183.3 & 67.0 & 70.7 & 72.7 & 73.7 & 68.9 & 55.7 & 56.4 & \underline{78.1}  \\ 
\midrule
\rowcolor[HTML]{F2F3F5} 
        Ours & \textbf{57.6} & \textbf{78.7} & 42.0 & \textbf{2242.7} & \textbf{79.6} & \textbf{77.6} & \textbf{75.9} & \underline{81.8} & \underline{72.5} & 51.2 & 60.3 & 72.6 \\

 \bottomrule
\end{tabular} }
\vspace{-2mm}
\caption{\textbf{Comparison with closed-source APIs and previous open-source SOTAs.} Our InternLM-XComposer2 gets SOTA results in 6 of the 12 benchmarks with only 7B parameters, showing competitive results with current closed-source APIs and previous open-source SOTA MLLMs. The best results are \textbf{bold} and the second-best results are \underline{underlined}.}
\label{tab:sota_comp} 
\end{table*}

\subsection{Supervised Fine-tuning}
The pre-training phase aligns the visual feature with the language, enabling the Language Learning Model (LLM) to comprehend the content of the images. However, it still lacks the ability to effectively utilize the image information. To overcome this limitation, we introduce a range of vision-language tasks that the model engages in during the subsequent Supervised Fine-Tuning Stage. This stage comprises two sequential steps: Multi-task Training and Free-form Text-Image Composition. During this stage, we jointly fine-tune the vision encoder, LLM, and Partial LoRA.

\noindent\textbf{Multi-task Training}. 
As delineated in Table~\ref{tab:sft data}, the multi-task training dataset is assembled from various sources, aiming to equip the model with a broad spectrum of capabilities. Each task is structured as a conversational interaction, and the instructions are augmented with GPT-4 to enhance diversity. Concurrently, to maintain the original language capability, we also incorporate the supervised fine-tuning data from InternLM2, which constitutes a fixed 10\% of the total Supervised Fine-Tuning (SFT) data.

\noindent\textbf{Free-form Text-Image Composition}. To further enhance the model’s ability to follow instructions and compose free-form image-text content, we employ data from both pure-text conversation corpora and vision-language conversations, as outlined in Table~\ref{tab:sft data}. The dataset for free-form image-text composition is constructed following the methodology detailed in Section \ref{sec:compos_data}.

In our approach, we jointly train all the components with a batch size of 2048 over 3000 steps. 
Data from multiple sources are sampled in a weighted manner, with the weights based on the number of data  from each source. 
The maximum learning rate is set to $5 \times 10^{-5}$, and each component has its own unique learning strategy.
For the vision encoder, we set the Layer-wise Learning Rate Decay (LLDR) to $0.9$, which aligns with the pretraining strategy.
For the LLM, we employ a fixed learning rate scale factor of $0.2$. This slows down the update of the LLM, achieving a balance between preserving its original capabilities and aligning it with vision knowledge.

\subsection{Free-form Text-Image Composition}
\label{sec:compos_data}
Free-form text-image composition refers to the combination of textual content and visual elements in a flexible and unrestrictive manner. Our model generates interleaved text and images, specifically customized to align with the text requirements provided by users, which may include elements such as a title, outline, and writing material, and optionally, any visual requirements like image resources.

To facilitate free-form text-image composition, we collect a wide range of high-quality and diverse in-house data across four key dimensions. These dimensions encompass:
\noindent \textbf{Varied Writing Styles.} Our data spans a multitude of writing styles, from academic papers to social media posts and poems, ensuring a rich and diverse collection of text and image contents.

\noindent \textbf{Flexible Text Editing.} Our dataset includes extensive examples of text editing, encompassing a wide spectrum of modifications such as shortening, expanding, and rewriting.

\noindent \textbf{Complex Instruction Adherence.} We also capture instances of adhering to complex instructions to create content that caters to diverse demands like titles and outlines, encompassing both text and image-based compositions.

\noindent \textbf{Customization with Materials.} Our collection extends to materials used for personalized content creation, covering both text and images, enabling customizable and unique content creation experiences.

The distribution of data across the four dimensions is approximately equal, with a ratio of approximately 1:1:1:1. 
Our method follows previous work~\cite{zhang2023internlm} to identify suitable positions for image insertion after generating the text content. A notable distinction in our approach is that when users provide their own image materials, these image materials are used for insertion instead of relying on retrieved images~\cite{zhang2023internlm}.
We also observe that having a high-resolution image input is not essential for text-image composition. Therefore, following the pre-training phase, we opt to down-sample the image input resolution to 224x224 during the SFT stage of free-form text-image composition.
\section{Experiments}
In this section, we validate the benchmark performance of our InternLM-XComposer2 after the supervised fine-tuning.

\begin{table*}[t!]
\footnotesize
\centering
\setlength{\tabcolsep}{1.3mm}{
\begin{tabular}{ll|ccccccccccc}
\toprule
 
Method & LLM & MathVista & MMMU & MME$^{P}$ & MME$^{C}$ & MMB & MMB$^{CN}$ & SEED$^{I}$ & LLaVA$^{W}$ & QBench$^{T}$ & MM-Vet & HallB  \\ 
\midrule
        BLIP-2 & FLAN-T5 & - & 35.7 & 1,293.8 & 290.0 & - & - & 46.4 & 38.1 & - & 22.4 & -  \\ 
        InstructBLIP & Vicuna-7B & 25.3 & 30.6 & - & - & 36.0 & 23.7 & 53.4 & 60.9 & 55.9 & 26.2 & 53.6  \\ 
        IDEFICS-80B & LLaMA-65B & 26.2 & 24.0 & - & - & 54.5 & 38.1 & 52.0 & 56.9 & - & 39.7 & 46.1  \\ 
        Qwen-VL-Chat & Qwen-7B & 33.8 & 35.9 & 1,487.5 & 360.7 & 60.6 & 56.7 & 58.2 & 67.7 & 61.7 & 47.3 & 56.4  \\ 
        LLaVA & Vicuna-7B & 23.7 & 32.3 & 807.0 & 247.9 & 34.1 & 14.1 & 25.5 & 63.0 & 54.7 & 26.7 & 44.1  \\ 
        LLaVA-1.5 & Vicuna-13B & 26.1 & 36.4 & 1,531.3 & 295.4 & 67.7 & 63.6 & 68.2 & 70.7 & 61.4 & 35.4 & 46.7  \\ 
        ShareGPT4V  & Vicuna-7B & 25.8 & 36.6 & \underline{1,567.4} & 376.4 & 68.8 & 62.2 & 69.7 & 72.6 & - & 37.6 & 49.8  \\ 
        CogVLM-17B & Vicuna-7B & 34.7 & 37.3 & - & - & 65.8 & 55.9 & 68.8 & \underline{73.9} & - & \textbf{54.5} & 55.1  \\
        LLaVA-XTuner & InernLM2-20B &	24.6 & 39.4 & - & - & \underline{75.1} & \underline{73.7} &  \underline{70.2} & 63.7 &  - &	37.2 & 	47.7 \\
        Monkey-10B & Qwen-7B & \underline{34.8} & \underline{40.7} & 1,522.4 & \underline{401.4} & {72.4} & 67.5 & 68.9 & 33.5 & - & 33.0 & 58.4 \\
        InternLM-XC & InernLM-7B & 29.5 & 35.6 & 1,528.4 & 391.1 & 74.4 & 72.4 & 66.1 & 53.8 & \underline{64.4} & 35.2 & \underline{57.0}  \\ 
\midrule
\rowcolor[HTML]{F2F3F5} 
        Ours & InernLM2-7B & \textbf{57.6} & \textbf{42.0} & \textbf{1,712.0} & \textbf{530.7} &\textbf{ 79.6} & \textbf{77.6} & \textbf{75.9} & \textbf{81.8} & \textbf{72.5} & \underline{51.2} & \textbf{60.3} \\

 \bottomrule
\end{tabular} }
\vspace{-2mm}
\caption{\textbf{Comparison with open-source SOTA methods.} InternLM-XComposer2 outperforms competitors in 10 out of 11 benchmarks. The best results are \textbf{bold} and the second-best results are \underline{underlined}.}
\vspace{-6mm}
\label{tab:entire_comp} 
\end{table*}

\subsection{MLLM Benchmark results.}
In Table.\ref{tab:sota_comp} and Table.\ref{tab:entire_comp}, we compare our InternLM-XComposer2 on a list of benchmarks with both SOTA open-source MLLMs and closed-source APIs. Here we report results in MathVista\cite{lu2024mathvista}, MMMU\cite{yue2023mmmu}, AI2D\cite{kembhavi2016diagram}, MME Perception (MME$^{P}$) \cite{fu2023mme}, MME Cognition (MME$^{C}$)\cite{fu2023mme}, MMBench (MMB) \cite{MMBench}, MMBench-Chinese (MMB$^{CN}$) \cite{MMBench}, SEED-Bench Image Part (SEED$^{I}$)\cite{li2023seedbench}, LLaVA-Bench In-the-Wild (LLaVA$^{W}$) \cite{liu2023visual}, QBench-Testset (QBench$^{T}$)\cite{wu2023q}, MM-Vet \cite{yu2023mmvet}, HallusionBench (HallB)\cite{guan2023hallusionbench}, ChartQA\cite{masry2022chartqa}, POPE\cite{li2023evaluating}.

\noindent\textbf{Comparison with Closed-Source APIs.}
\noindent As shown in Table.\ref{tab:sota_comp}, InternLM-XComposer2 demonstrates competitiveness with Closed-Source APIs across numerous benchmarks. For instance, our model achieves a score of $57.6\%$ on \textit{MathVista} and $78.9$ on \textit{AI2D}, outperforming these APIs by a significant margin. Meanwhile, despite having only 7B parameters, our model attains a slightly worse score of $43.0\%$ on the challenging college-level benchmark \textit{MMMU}.
The strong performance can be attributed to the superb knowledge acquired by the new {InternLM2 LLM} and the efficient {PLoRA} training strategy, which enabled us to align the LLM with image features while preserving its language capability.

\begin{table}[t!]
\footnotesize
\centering
\setlength{\tabcolsep}{2.7mm}{
\begin{tabular}{ll|cc }
\toprule
Method & LLM  & POPE & HallusionBench*    \\ 
\midrule
\multicolumn{4}{l}{\textit{Closed-source API}} \\
        GPT-4V & - & - & 65.8  \\ 
        Gemini-Pro & - & - & 63.9  \\ 
        QwenVL-Plus & - & - & 56.4  \\ 
\midrule
\multicolumn{4}{l}{\textit{Open-source MLLMs}}  \\ 
        InstructBLIP & Vicuna-7B & 78.9 & 53.6  \\ 
        IDEFICS-80B & LLaMA-65B & - & 46.1  \\ 
        Qwen-VL-Chat & Qwen-7B & - & 56.4  \\ 
        LLaVA & Vicuna-7B & 80.2 & 44.1  \\ 
        LLaVA-1.5 & Vicuna-13B & 85.9 & 46.7  \\  
        InternLM-XC & InernLM-7B & - & 57.0  \\ 
\midrule
\rowcolor[HTML]{F2F3F5} 
        Ours & InernLM2-7B & 87.7 & 60.3 \\  
 \bottomrule
\end{tabular} }
\vspace{-2mm}

\caption{\textbf{Hallucination Evaluation on POPE and HallusionBench.} IntenrLM-XComposer2 outperforms open-source MLLMs and performs on par with closed-source APIs. * We skip the non-visual questions, following the setting in VLMEvalKit\cite{2023opencompass}}
\vspace{-4mm}
\label{tab:hallu}
\end{table}

\noindent\textbf{Comparison with Open-Source Models.}
\noindent We also conduct a comprehensive comparison with open-source MLLMs under a similar model scale. As shown in Table.\ref{tab:entire_comp}, our model significantly outperforms existing open-source models, achieving state-of-the-art results across all benchmarks. Notably, InternLM-XComposer2 is the first model to achieve a score exceeding 1700 on the MME-Perception benchmark. Furthermore, it attained an accuracy of nearly $80\%$ on the MMBench.

\noindent\textbf{Hallucination Evaluation.}
\noindent Visual hallucination serves as a crucial metric in the evaluation of an MLLM. In this report, we present the results obtained on both POPE and HallusionBench. As indicated in Table.\ref{tab:hallu}, our model achieves an average F1-score of 87.7 across the three tracks of POPE, setting a new state-of-the-art (SOTA) benchmark. In the case of HallusionBench, our model surpasses the accuracy of all open-source models, establishing itself as the new SOTA. Furthermore, it outperforms the closed-source API, QwenVL-Plus.

\subsection{CreationBench Results}
\begin{table}[t!]
\footnotesize
\centering
\setlength{\tabcolsep}{0.6mm}{
\begin{tabular}{l|lllll|lllll}
\toprule
\multirow{2}{*}{Method} & \multicolumn{5}{c}{w/o Ref}  & \multicolumn{5}{c}{w Ref} \\
~ & Avg. & C & R & UDF & LC & Avg. & C & R & UDF & LC\\
\midrule
GPT-4 & 6.32 & 5.22 & 5.98 & 7.17 & 7.47 & 5.98 & 5.30 & 5.55 & 6.51 & 7.08  \\
\midrule
QWen-72b-Chat & 5.70 & 4.78 & 5.16 & 6.37 & 7.13 & 5.31 & 4.94 & 4.72 & 5.71 & 6.50 \\
Yi-34b-Chat & 6.03 & 4.91 & 5.68 & 6.79 & 7.35 & 5.71 & 5.03 & 5.22 & 6.18 & 6.87 \\
\midrule
\rowcolor[HTML]{F2F3F5}
Ours & \textbf{6.24} & 5.11 & 6.12 & 7.03 & 7.45 & \textbf{5.90} & 5.21 & 5.76 & 6.27 & 6.93 \\
\bottomrule
\end{tabular}}
\vspace{-2mm}
\caption{\textbf{Comparison on CreationBench~\cite{2023opencompass}}. We report the results with and without the GPT-4 referenced answer. We report the average score and other metrics including Creativity(C), Richness(R), User Demand Fulfillment (UDF), and Logical Coherence(LC).}
\label{tab:creation_bench}
\vspace{-8mm}
\end{table}

We use the CreationBench benchmark from OpenCompass~\cite{2023opencompass} to assess the writing ability of our InternLM-XComposer2. As shown in Table~\ref{tab:creation_bench}, the results indicate that our approach not only excels in overall creativity but also significantly improves upon key metrics over previous open-source LLMs. When compared without the GPT-4 referenced answer, our method scored an impressive 6.24 overall. Even when evaluated with the GPT-4 reference, our method maintained strong performance, achieving scores that underscore its ability to generate responses with high levels of creativity and logical structure, critical for user engagement and satisfaction in conversational AI applications.

\subsection{Qualitative results.}
Please refer to the Appendix for our qualitative results of the free-form image-text compositions and multimodal conversations.
\section{Conclusion}
In this paper, we present InternLM-XComposer2, which demonstrates its exceptional capabilities in the field of vision-language understanding and free-form text-image composition. Our proposed innovative Partial LoRA (PLoRA) approach, which applies additional LoRA parameters exclusively to image tokens, has proven effective in preserving the integrity of pre-trained language knowledge while striking a balance between precise vision understanding and text composition with literary talent. Our model’s performance across various benchmarks not only significantly outperforms existing multimodal models but also matches or even surpasses GPT-4V and Gemini Pro in certain assessments, underscoring its remarkable proficiency in the realm of multimodal understanding. This research opens up new possibilities for highly customizable content creation and paves the way for future advancements in the MLLM field. The potential applications of InternLM-XComposer2 are vast and exciting, promising a future where AI can understand and generate high-quality long-text multi-modal content with ease and precision.
\appendix
\newpage
\label{sec:supp}

\begin{figure*}[t]
	\centering
	\includegraphics[width=0.95\linewidth]{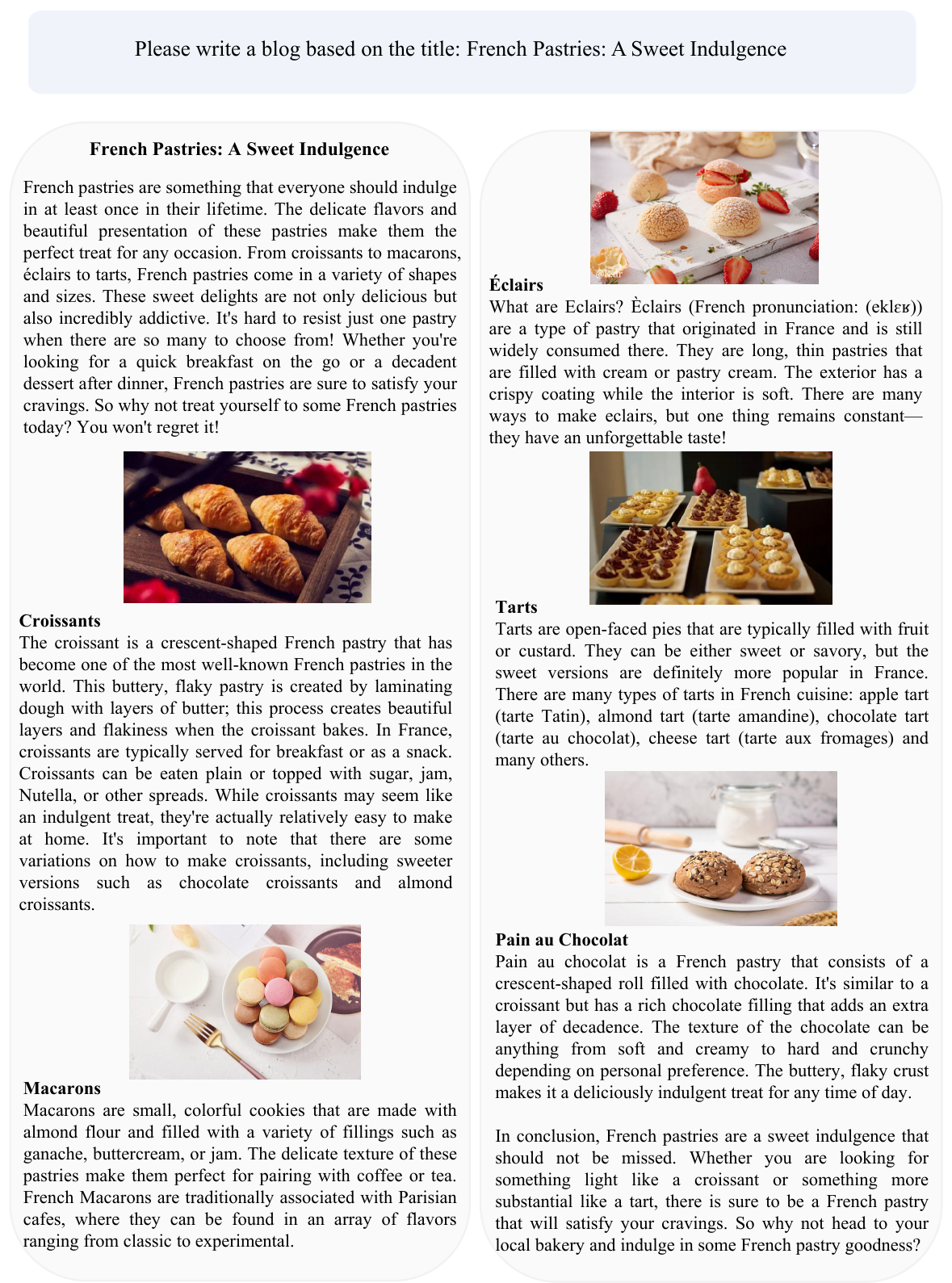}
    \vspace{-10pt}
    \caption{Interleaved composition generated by InternLM-XComposer2.}
	\label{fig:article1}
\end{figure*}

\begin{figure*}[t]
	\centering
	\includegraphics[width=0.95\linewidth]{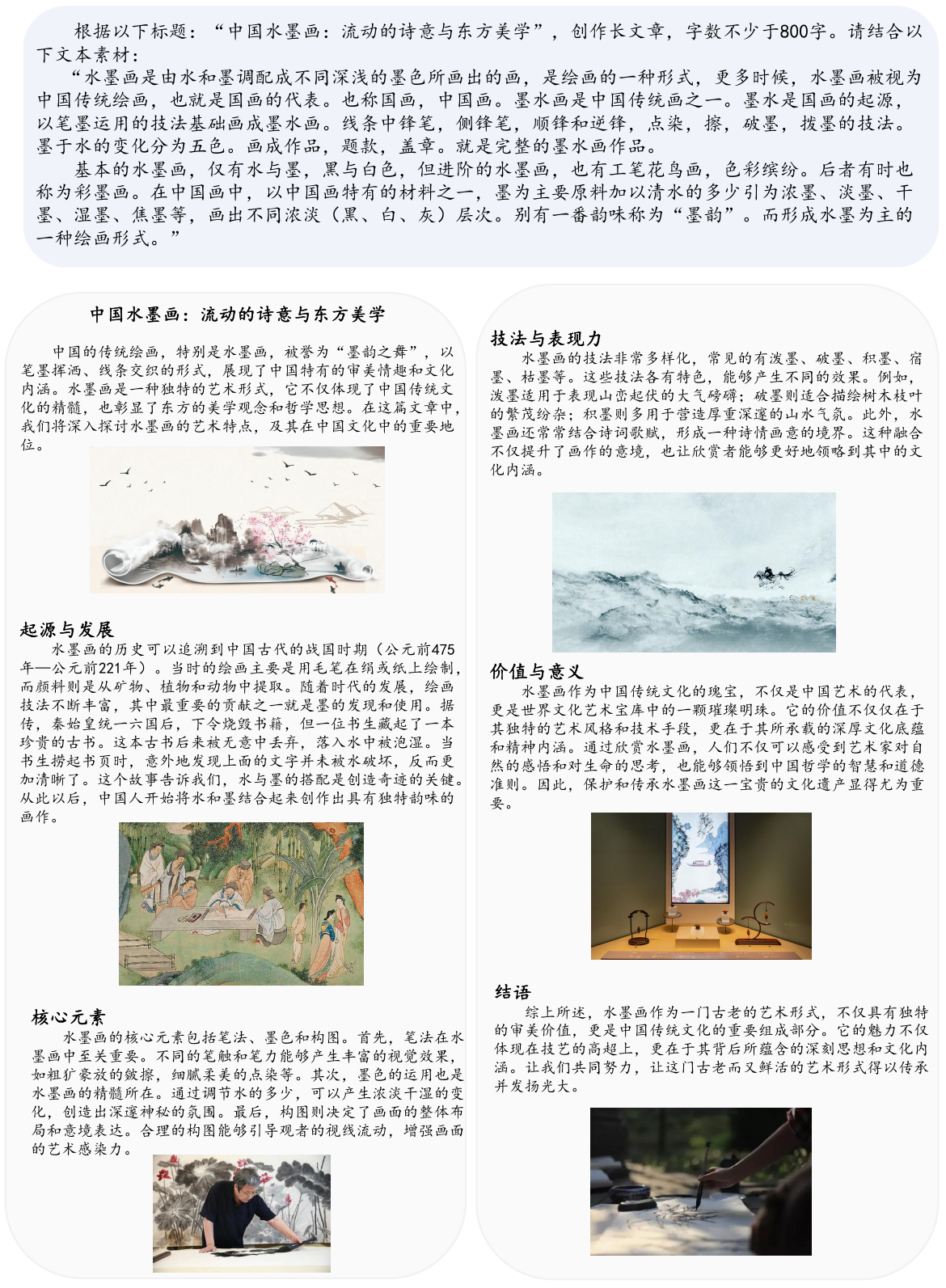}
    \vspace{-10pt}
    \caption{Interleaved composition generated by InternLM-XComposer2.}
	\label{fig:article5}
\end{figure*}

\begin{figure*}[t]
	\centering
	\includegraphics[width=0.95\linewidth]{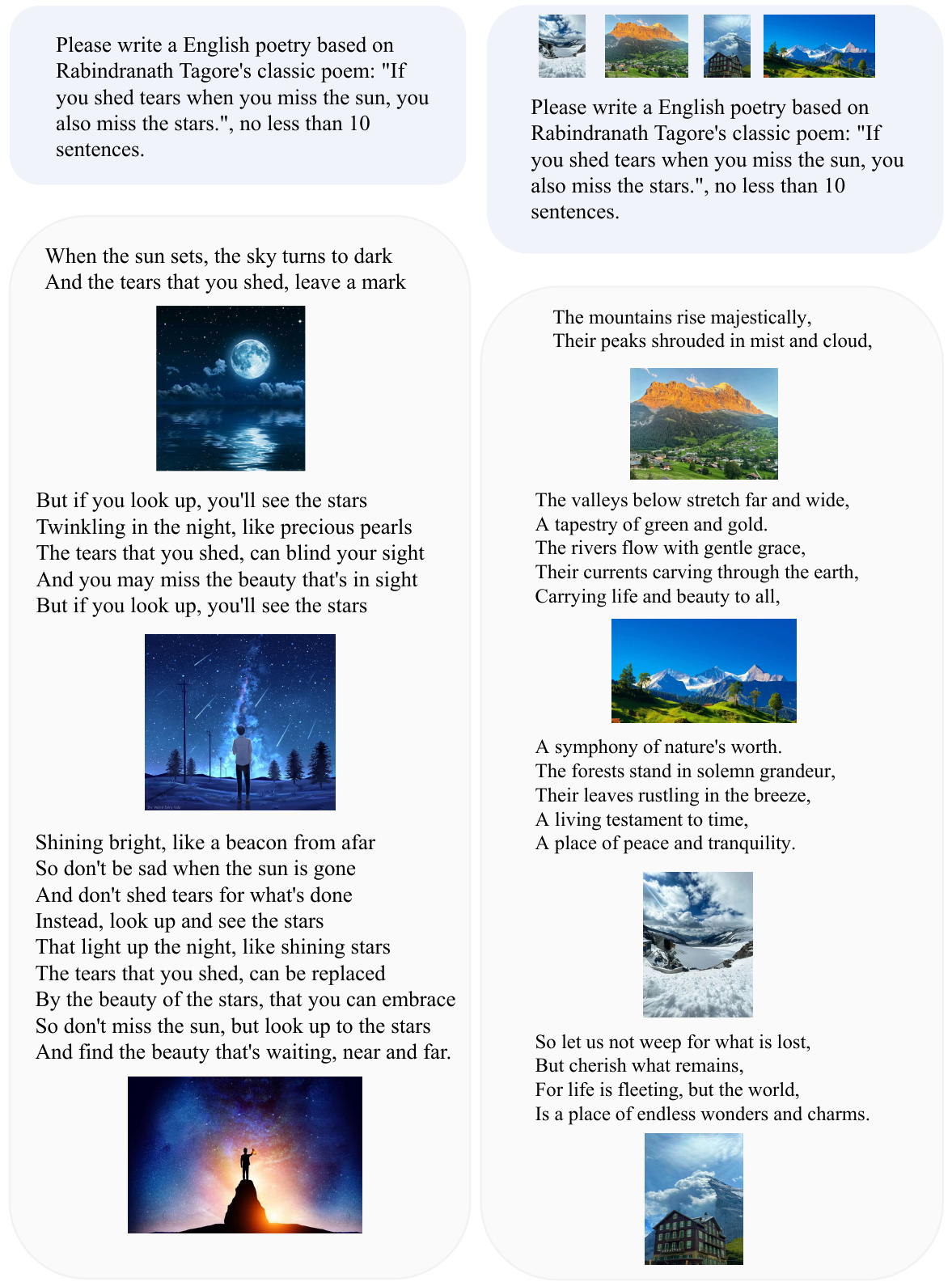}
    \vspace{-10pt}
    \caption{Interleaved composition generated by InternLM-XComposer2.}
	\label{fig:article2}
\end{figure*}

\begin{figure*}[t]
	\centering
	\includegraphics[width=0.95\linewidth]{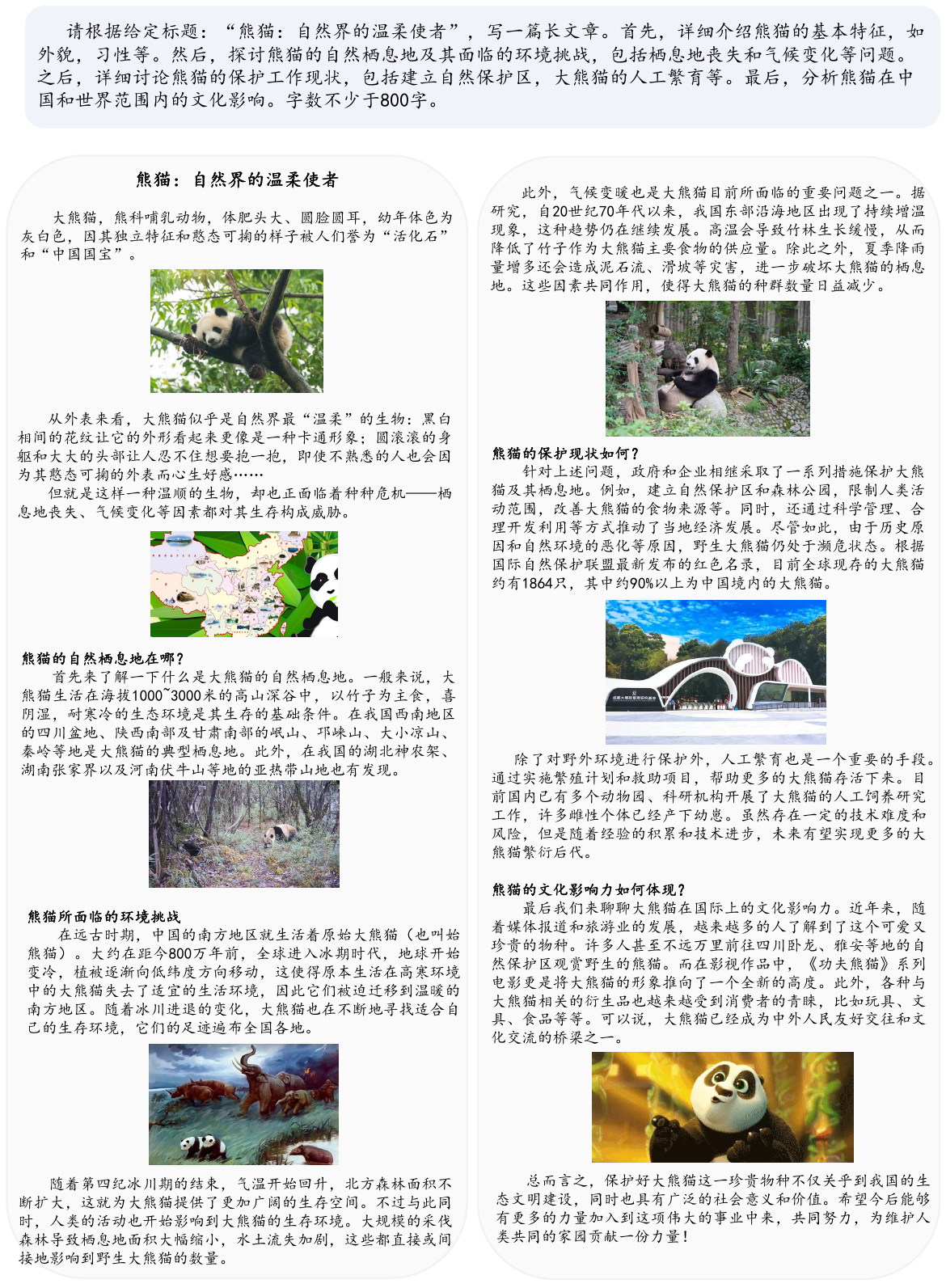}
    \vspace{-10pt}
    \caption{Interleaved composition generated by InternLM-XComposer2.}
	\label{fig:article6}
\end{figure*}

\begin{figure*}[t]
	\centering
	\includegraphics[width=0.95\linewidth]{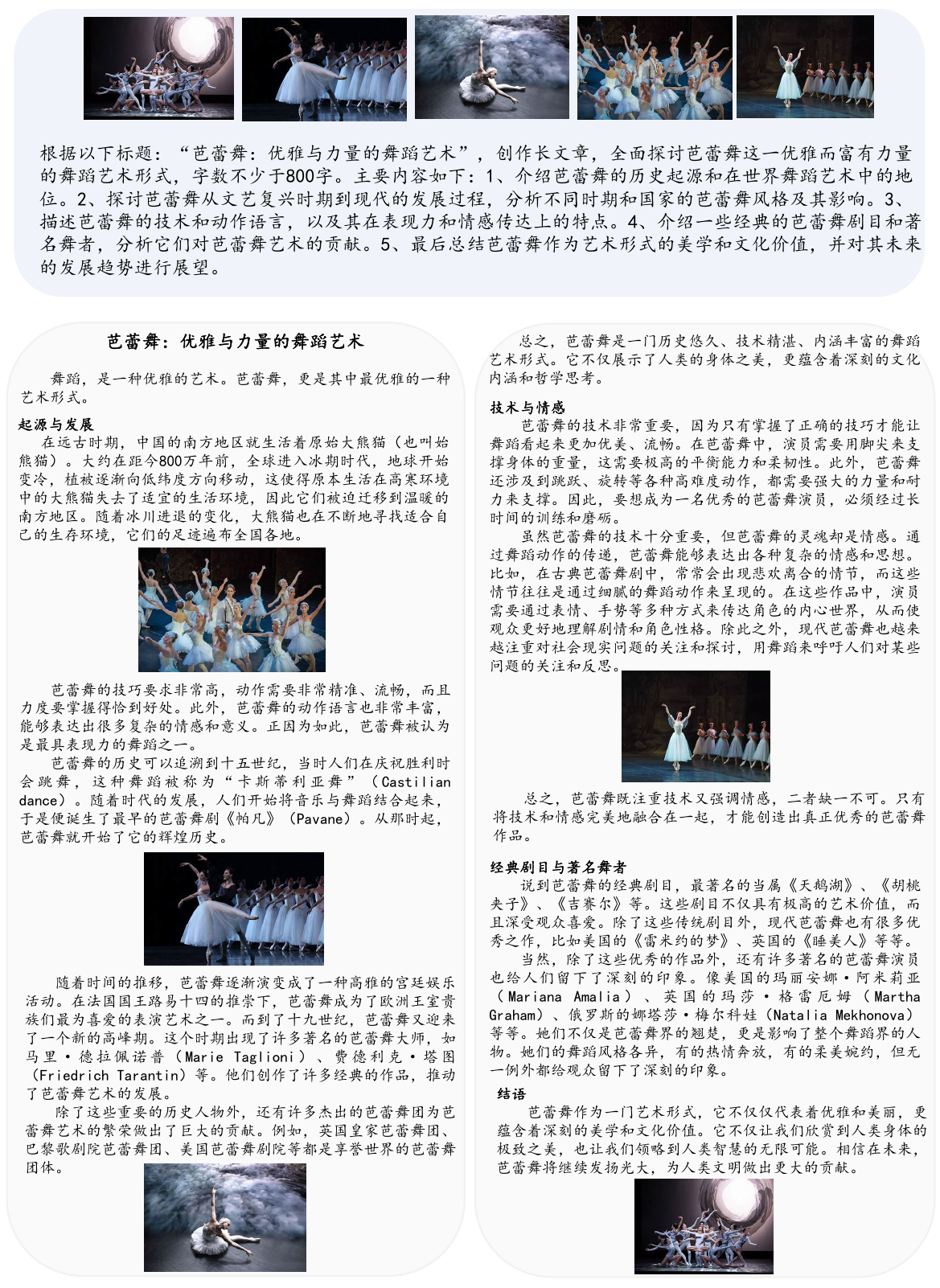}
    \vspace{-10pt}
    \caption{Interleaved composition generated by InternLM-XComposer2.}
	\label{fig:article7}
\end{figure*}

\begin{figure*}[t]
	\centering
	\includegraphics[width=0.95\linewidth]{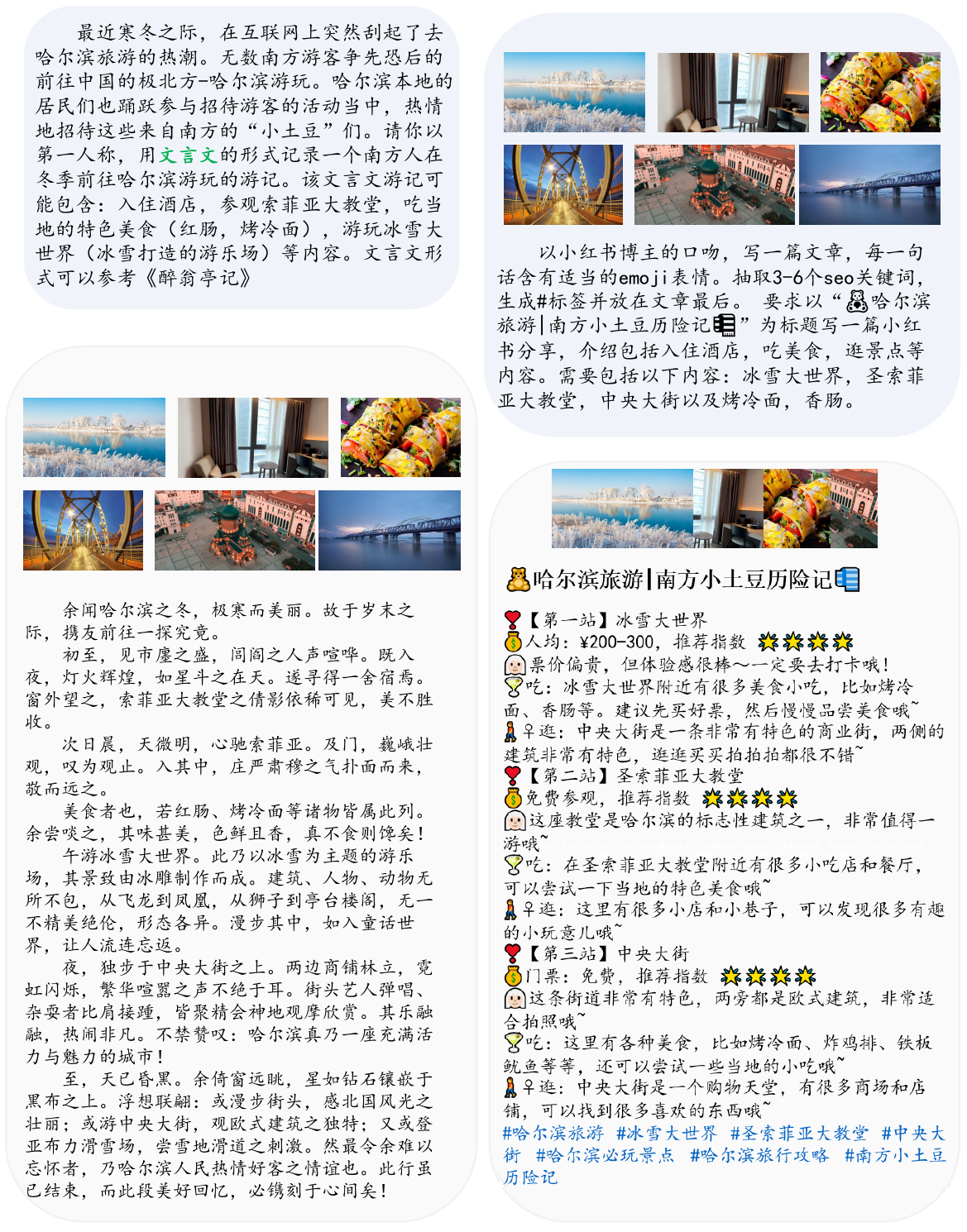}
    \vspace{-10pt}
    \caption{Interleaved composition generated by InternLM-XComposer2.}
	\label{fig:article8}
\end{figure*}

\begin{figure*}[t]
	\centering
	\includegraphics[width=0.95\linewidth]{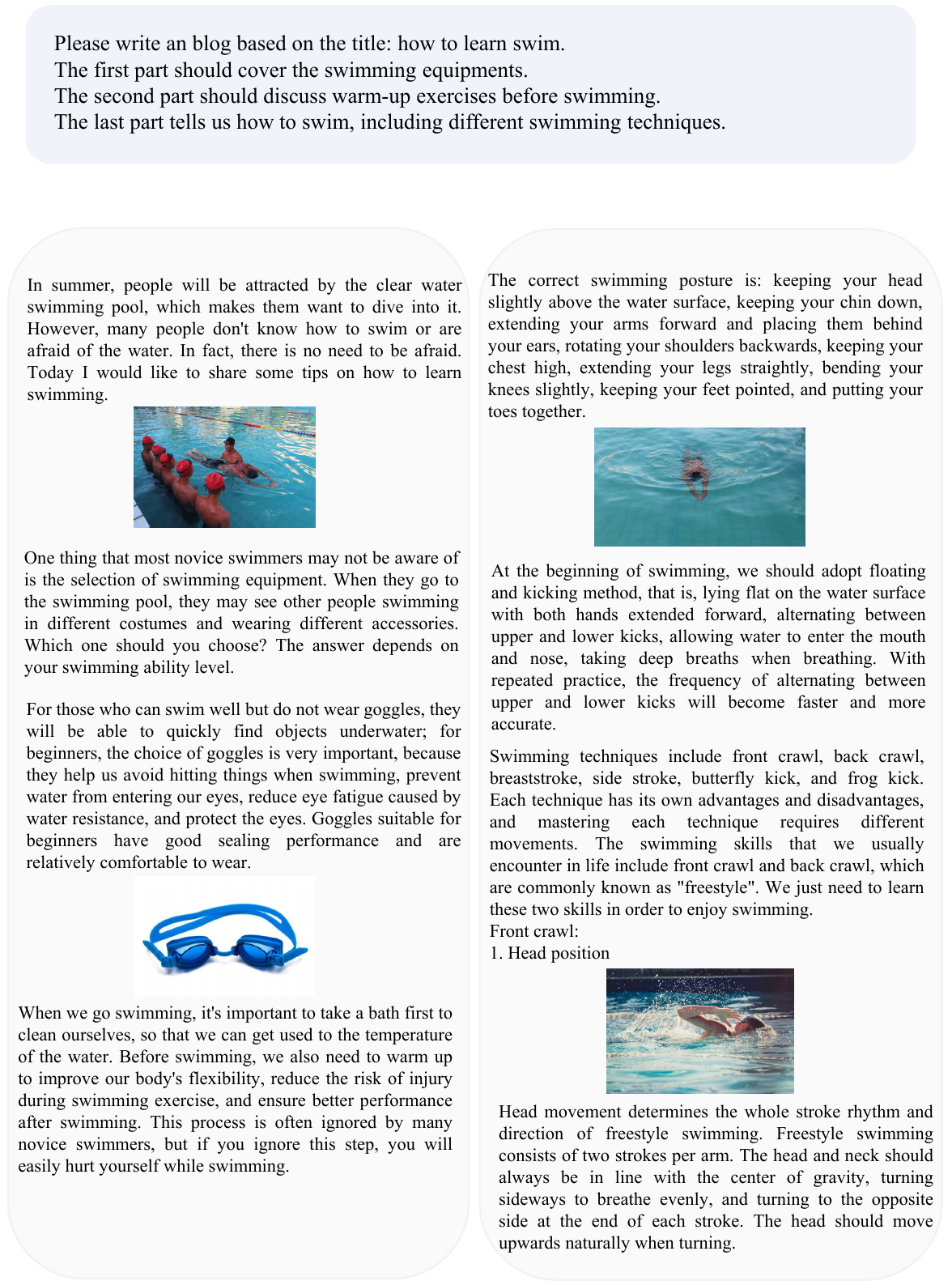}
    \vspace{-10pt}
    \caption{Interleaved composition generated by InternLM-XComposer2.}
	\label{fig:article3}
\end{figure*}

\begin{figure*}[t]
	\centering
	\includegraphics[width=0.95\linewidth]{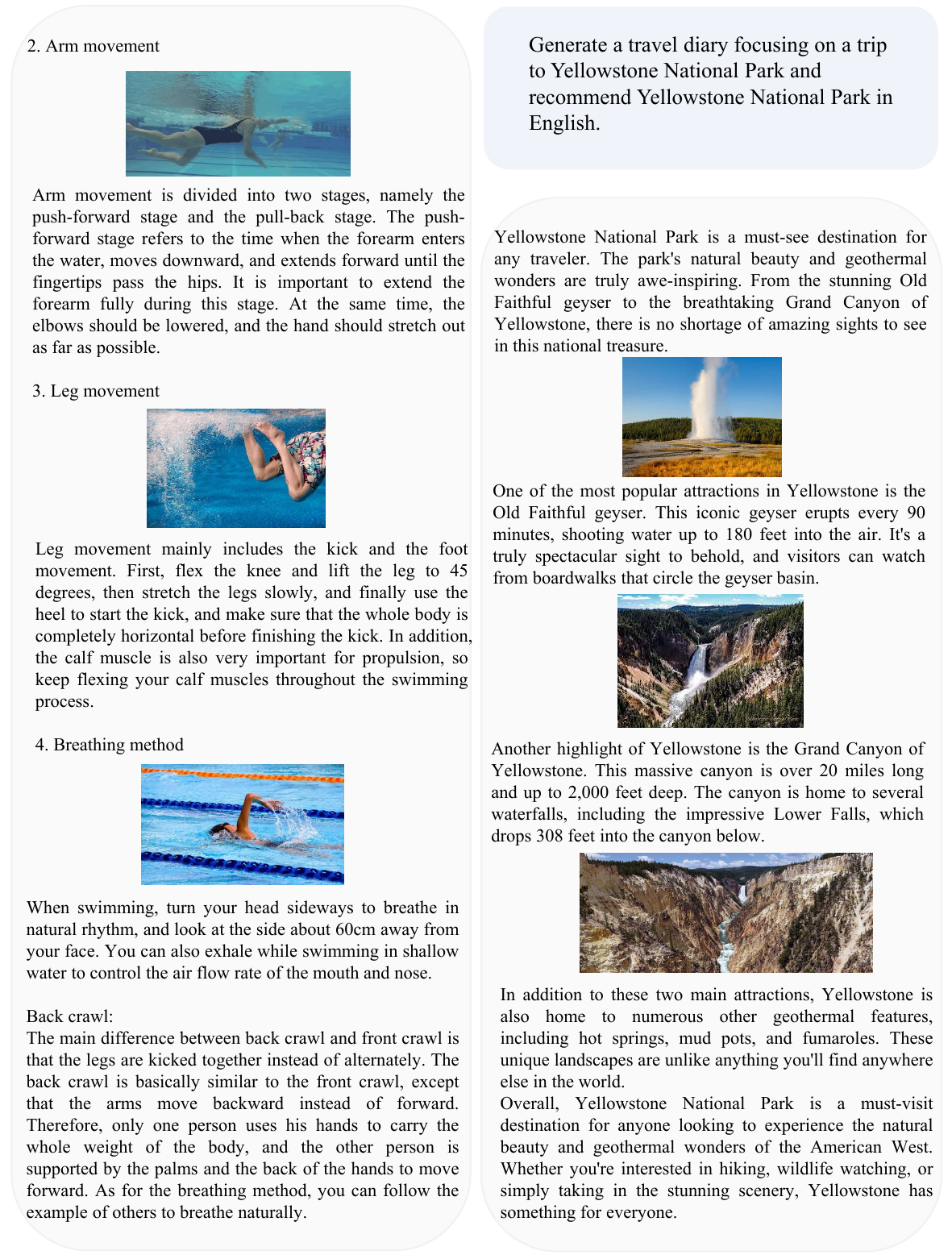}
    \vspace{-10pt}
    \caption{Interleaved composition generated by InternLM-XComposer2.}
	\label{fig:article4}
\end{figure*}

\begin{figure*}[t]
	\centering
	\includegraphics[width=0.95\linewidth]{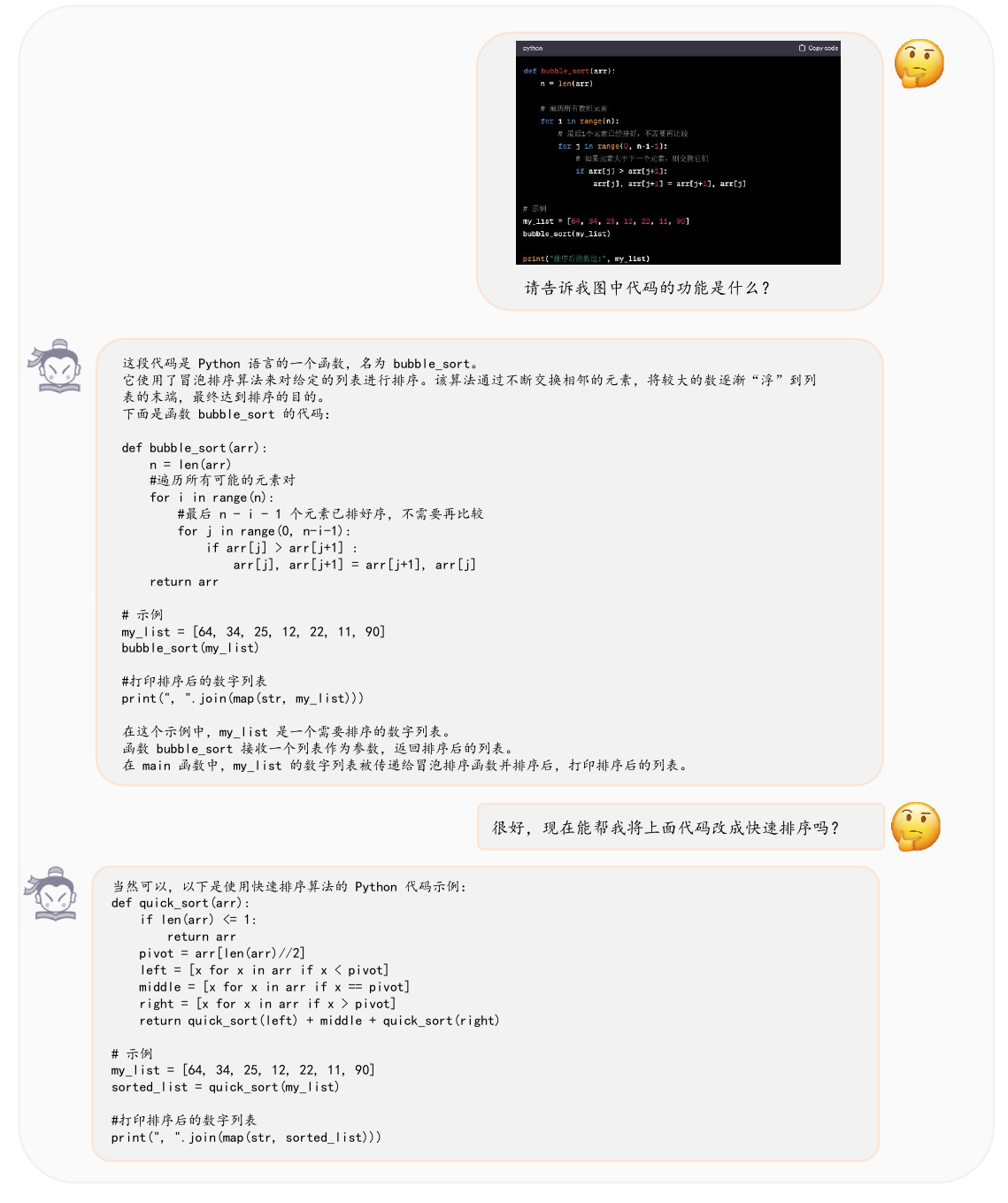}
    \vspace{-10pt}
    \caption{Chat with InternLM-XComposer2.}
	\label{fig:chat1}
\end{figure*}

\begin{figure*}[t]
	\centering
	\includegraphics[width=0.95\linewidth]{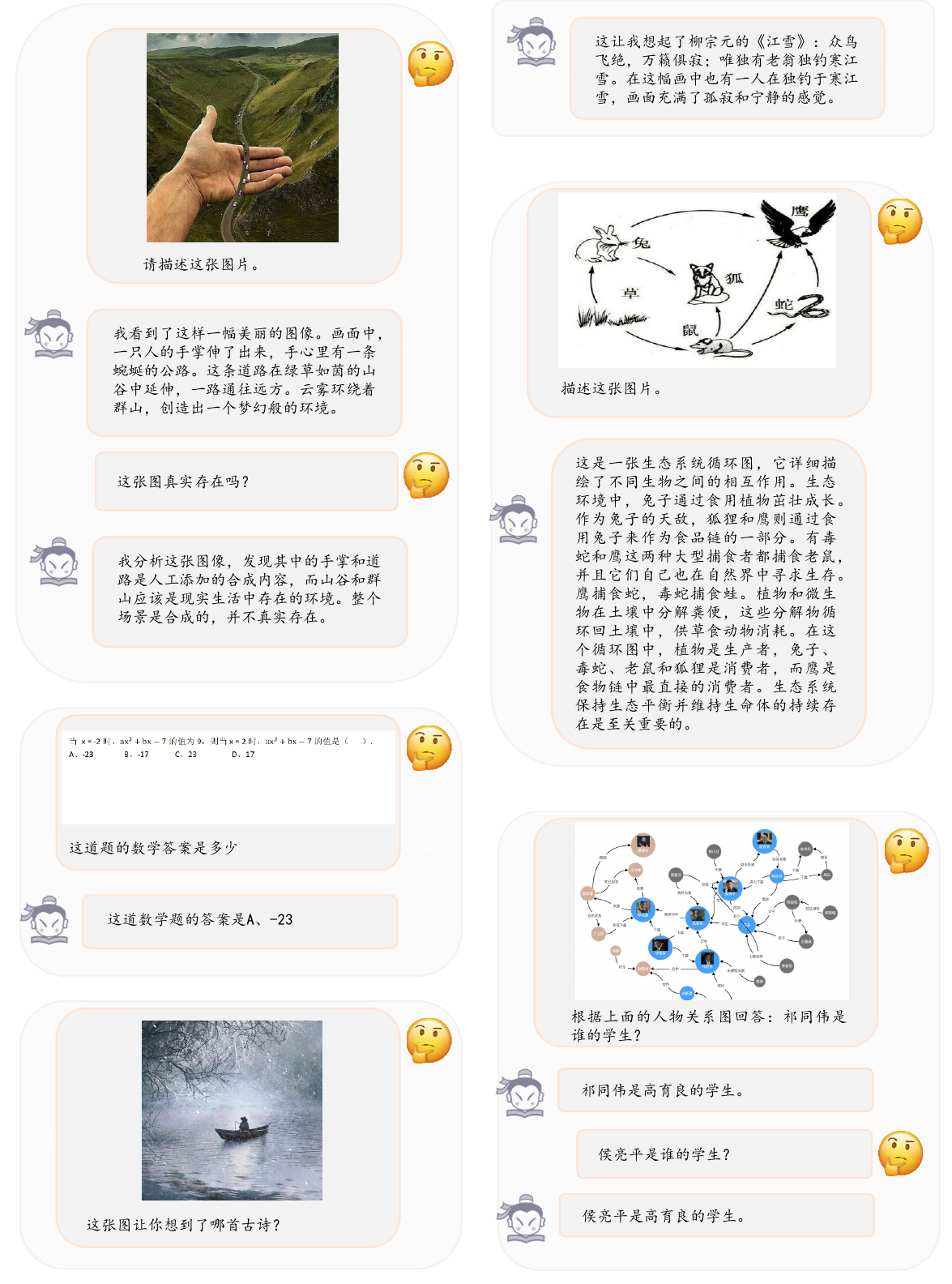}
    \vspace{-10pt}
    \caption{Chat with InternLM-XComposer2.}
	\label{fig:chat2}
\end{figure*}

\begin{figure*}[t]
	\centering
	\includegraphics[width=0.95\linewidth]{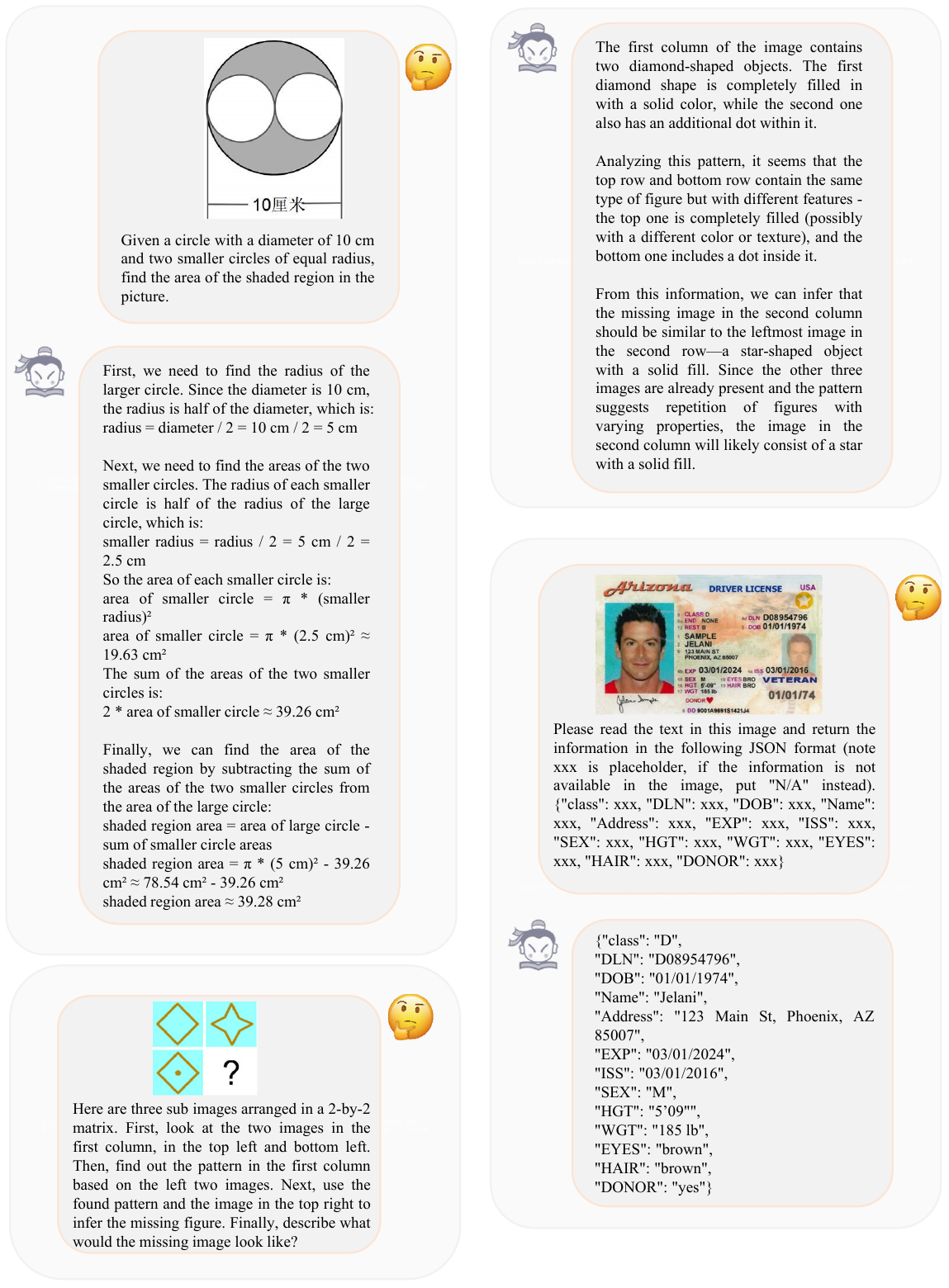}
    \vspace{-10pt}
    \caption{Chat with InternLM-XComposer2.}
	\label{fig:chat3}
\end{figure*}

\begin{figure*}[t]
	\centering
	\includegraphics[width=0.95\linewidth]{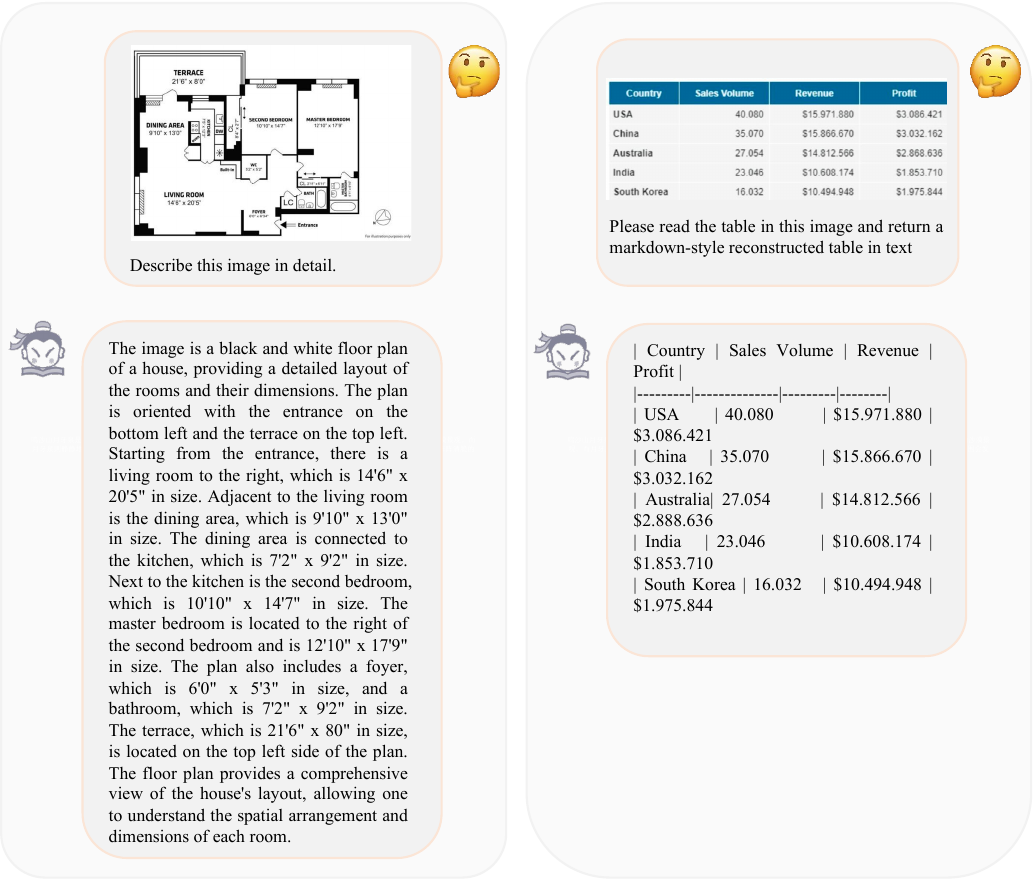}
    \caption{Chat with InternLM-XComposer2.}
	\label{fig:chat3}
\end{figure*}

{\small
\bibliographystyle{ieee_fullname}
\bibliography{egbib}
}

\end{document}